# Evaluating the Robustness of Learning from Implicit Feedback

**Filip Radlinski**                                                                   FILIP@CS.CORNELL.EDU

Department of Computer Science, Cornell University, Ithaca, NY 14853 USA

**Thorsten Joachims**                                                                 TJ@CS.CORNELL.EDU

Department of Computer Science, Cornell University, Ithaca, NY 14853 USA

## Abstract

This paper evaluates the robustness of learning from implicit feedback in web search. In particular, we create a model of user behavior by drawing upon user studies in laboratory and real-world settings. The model is used to understand the effect of user behavior on the performance of a learning algorithm for ranked retrieval. We explore a wide range of possible user behaviors and find that learning from implicit feedback can be surprisingly robust. This complements previous results that demonstrated our algorithm's effectiveness in a real-world search engine application.

## 1. Introduction

The task of learning ranked retrieval functions has recently received significant interest in the machine learning community (Bartell & Cottrell, 1995; Freund et al., 1998; Joachims, 2002; Kemp & Ramamohanarao, 2003). This is largely motivated by a goal of learning improved retrieval functions for web search.

The two standard approaches for collecting training data in this setting use explicit and implicit feedback. Explicit feedback involves actively soliciting relevance feedback by recording user queries and then explicitly judging the relevance of the results (Crammer & Singer, 2001; Herbrich et al., 2000; Rajaram et al., 2003). Acquiring explicit relevance judgments is time consuming and tedious, making large amounts of such data impractical to obtain. The alternative is to extract implicit relevance feedback from search engine log files (Kelly & Teevan, 2003; Cohen et al., 1999; Joachims, 2002; Kemp & Ramamohanarao, 2003). This allows virtually unlimited data to be collected at very low cost, but this data tends to be noisy and biased (Joachims et al., 2005; Radlinski & Joachims, 2005). In this paper, we consider a method for learning from implicit feedback and use modeling to understand when it is effective.

In contrast to typical learning problems where we have a fixed dataset, the task of learning to rank from implicit feedback is an interactive process between the user and learning algorithm. The training data is collected by observing user behavior given a particular ranking. If an algorithm presents users with a different ranking, different training data will be collected.

This type of interactive learning requires that we either run systems with real users, or build simulations to evaluate algorithm performance. The first involves building a search system to collect training data and evaluate real user behavior. While providing the most compelling results, this approach has a number of drawbacks. First, evaluating with real users is slow and requires a significant number of different users. Moreover, if a particular learning method proves ineffective, users quickly switch to other search engines. Finally, when we only collect the behavior of real users, the behavior is determined by the user base. Such results do not allow us to study the robustness of learning algorithms and feedback mechanisms. It is this issue that is our primary concern in this paper.

The alternative, often used in reinforcement learning, is to build a simulation environment. Obviously this has the drawback that it is merely a simulation, but it also has significant advantages. It allows more rapid testing of algorithms than by relying on user participation. It also allows exploration of the parameters of user behavior. In particular, we can use a model to explore the robustness of a learning algorithm to noise in the training data. We cannot have such control when real users are involved, and unlike the usual learning problem setting we are unaware of any way to inject realistic implicit feedback noise into real-world training data and evaluate its effect.

In this paper, we present a user model to analyze the robustness of the Osmot search engine (Radlinski & Joachims, 2005). Osmot learns ranked retrieval functions by observing how users reformulate queries and how they click on results. We first present the learning algorithm, then the user model where we draw on the results of an eye-tracking study (Granka et al., 2004). We next demonstrate our algorithm's tolerance to noise in user behavior, having previously shown it to be effective in a real-world search engine (Radlinski & Joachims, 2005). We find Osmot to tolerate a strong user preference to click on higher ranked documents, and that it is able to learn despite most users only looking at the top few results. Our approach is generally interesting because it provides a practical method to evaluate the robustness of learning from implicit feedback. We plan to publicly release Osmot, including our model implementation.

## 2. Learning to Rank

Before we present our simulation model, we describe how Osmot learns from implicit feedback. For this, we assume a standard web search setting.

Our method relies on implicit feedback collected from log files. We record the queries users run as well as the documents they click on in the results. In these log files, we assume that documents clicked on are likely more relevant than documents seen earlier by the user, but not clicked on. This allows us to extract implicit relevance judgments according to a given set of feedback strategies. Within each search session, we assume each user runs a sequence, or chain, of queries while looking for information on some topic. We segment the log data into query chains using a simple heuristic (Radlinski & Joachims, 2005).

### 2.1. Implicit Feedback Strategies

We generate preference feedback using the six strategies illustrated in Figure 1. They are validated and discussed more in (Radlinski & Joachims, 2005). The first two strategies show single query preferences. "Click $>_q$ Skip Above" proposes that given a clicked-on document, any higher ranked document that was not clicked on is less relevant. The preference is indicated by an arrow labeled with the query, to show that the preference is with respect to that query. Note that these preferences are not stating that the clicked-on document *is* relevant, rather that it is *more likely* to be relevant than the ones not clicked on. The second strategy, "Click $1^{st} >_q$ No-Click $2^{nd}$" assumes that users typically view both of the top two results be-fore clicking, as suggested by an eye-tracking study described below (Joachims et al., 2005). It states that if the first document is clicked on, but the second is not, the first is likely more relevant than the second.

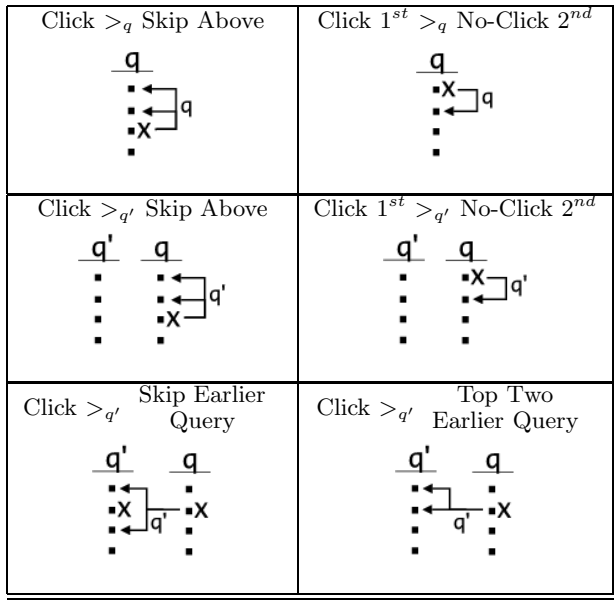

Figure 1. Feedback strategies. The user has run query $q'$ followed by $q$. Each dot represents a result and an $x$ indicates it was clicked on. We generate a constraint for each arrow shown, with respect to the query marked.

The next two strategies are identical to the first two except that they generate feedback with respect to the *earlier* query. The intuition is that since the two queries belong to the same query chain, the user is looking for the same information with both. Had the user been presented with the later results for the earlier query, she would have preferred the clicked-on document over those skipped over.

The last two strategies make the most use of query chains. They state that a clicked-on result is preferred over any result not clicked on in an earlier query (within the same query chain). This judgment is made with respect to the earlier query. We assume the user looked at all the documents in the earlier query up to one past the last one clicked on. In the event that no documents were clicked on in the earlier query, we assume the user looked at the top two results.

Ultimately, given some query chain, we make use of all six strategies as illustrated in the example in Figure 2.

### 2.2. Learning ranking functions

We define the relevance of $d_i$ to $q$ as a linear function,

$$rel(d_i, q) := w \cdot \Phi(d_i, q) \quad (1)$$

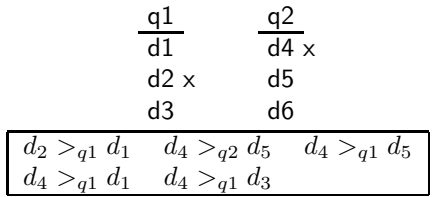

*Figure 2.* Sample query chain and the feedback that would be generated. Two queries were run, each returning three results of which one was clicked on. $d_i >_q d_j$ means that $d_i$ is preferred over $d_j$ with respect to the query $q$.

where $\Phi(d_i, q)$ maps documents and queries to a feature vector. Intuitively, $\Phi$ can be though of as describing the quality of the match between $d_i$ and the query $q$. $w$ is a weight vector that assigns weights to each of the features in $\Phi$, giving a real valued retrieval function where a higher score indicates $d_i$ is estimated to be more relevant to $q$. The task of learning a ranking function becomes one of learning $w$.

The definition of $\Phi(d_i, q)$ is key in determining the class of ranking functions we can learn. We define two types of features: rank features, $\phi_{rank}(d_i, q)$, and term/document features, $\phi_{terms}(d_i, q)$. Rank features serve to exploit an existing static retrieval function $rel_0$, while term/document features allow us to learn fine-grained relationships between particular query terms and specific documents. Note that $rel_0$ is the only ranking function we have before any learning has occurred and is thus used to generate the original ranking of documents. In our case, we use a simple TFIDF weighted cosine similarity metric as $rel_0$.

Let $W := \{t_1, \ldots, t_N\}$ be all the terms in our dictionary. A query $q$ is a set of terms $q := \{t'_1, \ldots, t'_n\}$ where $t'_i \in W$. Let $D := \{d_1, \ldots, d_M\}$ be the set of all documents. We also define $r_0(q)$ as the ordered set of results as ranked by $rel_0$ given query $q$. Now,

$$\Phi(d, q) = \left[ \begin{array}{c} \phi_{rank}(d, q) \\ \phi_{terms}(d, q) \end{array} \right]$$

$$\phi_{rank}(d, q) = \left[ \begin{array}{c} \mathbf{1}(Rank(d \ in \ r_0(q)) \leq 1) \\ \vdots \\ \mathbf{1}(Rank(d \ in \ r_0(q)) \leq 100) \end{array} \right]$$

$$\phi_{terms}(d, q) = \left[ \begin{array}{c} \mathbf{1}(d = d_1 \wedge t_1 \in q) \\ \vdots \\ \mathbf{1}(d = d_M \wedge t_N \in q) \end{array} \right]$$

where $\mathbf{1}$ is the indicator function.

Before looking at the term features $\phi_{terms}(d, q)$, consider the rank features $\phi_{rank}(d, q)$. We have 28 rank features (for ranks 1,2,..,10,15,20,..,100), with each set to 1 if document $d$ in $r_0(q)$ is at or above the specified rank. The rank features allow us to make use of the original ranking function.

The term features, $\phi_{terms}(d, q)$, are each of the form $\phi_{term}^{t_i, d_j}(d, q)$, set to either 0 or 1. There is one for every (term, document) pair in $W \times D$. These features allow the ranking function to learn associations between specific query words and documents. This is usually a very large number of features, although most never appear in the training data. Furthermore, the feature vector $\phi_{terms}(d, q)$ is very sparse. For any particular document $d$, given a query with $|q|$ terms, only $|q|$ of the $\phi_{term}^{t_i, d_j}(d, q)$ features are set to 1.

We use a modified ranking SVM (Joachims, 2002) to learn $w$ from Equation 1. Let $d_i$ be more relevant than $d_j$ to query $q$: $rel(d_i, q) > rel(d_j, q)$. We can rewrite this, adding margin and non-negative slack variables:

$$w \cdot \Phi(d_i, \ q) \geq w \cdot \Phi(d_j, \ q) + 1 - \xi_{ij} \qquad (2)$$

We also have additional prior knowledge that absent any other information, documents with a higher rank in $r_0(q)$ should be ranked higher in the learned ranking system. There are both intuitive and practical reasons for these constraints (Radlinski & Joachims, 2005).

This gives the following optimization problem that we solve using $SVM^{light}$ (Joachims, 1999) with $C = 0.1$:

$$\begin{array}{l} \min_{w, \xi_{ij}} \frac{1}{2} w \cdot w + C \sum_{ij} \xi_{ij} \text{ subject to} \\ \forall (q, i, j) : \ w \cdot (\Phi(d_i, q) - \Phi(d_j, q)) \geq 1 - \xi_{ij} \\ \forall i \in [1, \ 28] : \ w^i \geq 0.01 \\ \forall i, j : \ \xi_{ij} \geq 0 \end{array} \qquad (3)$$

We have shown that this algorithm works in a real-world setting in the Cornell University library web search engine (Radlinski & Joachims, 2005). Due to space constraints we do not repeat those results here.

## 3. Model Description

We now present a model of user behavior when searching. This model will allow us to measure the robustness of Osmot to changes in user behavior. One part generates documents, and another simulates users searching the collection. After presenting the model, we support it by drawing on user behavior studies. Although it is clearly a simplification of reality, we show that this model is nonetheless useful.

### 3.1. Document Generation

Documents are generated as described in Table 1. The set of words is $W$, with word frequencies obeying a Zipf law. We define a set of topics $T$ by uniformly picking $N$ words from $W$ for each topic. Some topics thus include

Evaluating the Robustness of Learning from Implicit Feedback

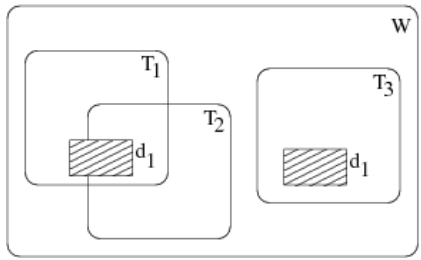

*Figure 3.* Document generation illustration. $T_1$, $T_2$ and $T_3$ are topics. Document $d_1$ is picked as relevant to two topics ($k_d = 2$), $T_1$ and $T_3$, although in selecting words from $T_1$, we also happened to select some words in $T_2$.

*Table 1.* Document Generation Model.

1. Let $W$ be the set of all words. Let $T$ be the set of topics, with each topic described by $T_i \subset W$.

2. Let each document $d$ be generated as follows:

   2.1. $\forall T_i \in T : rel(d, T_i) = 0$
   2.2. Pick $k_d$ binomially from $[0, MAX_T]$.
   2.3. If $k_d = 0$ Then
      Pick $L$ words from $W$.
   2.4. Otherwise, do the following $k_d$ times
      a. Pick $t$ from $[1, |T|]$.
      b. Pick $L/k_d$ words from $T_t$.
      c. $rel(d, T_t) = rel(d, T_t) + 1/k_d$.

*Table 2.* User Behavior Model

1. Let $q$ be the user's question, and $p$ and $r$ the user's patience and relevance thresholds respectively. They are sampled uniformly from $(0,5]$ and $[0.375, 0.875]$ respectively.

2. While question $q$ is unanswered

   2.1. Generate a query for question $q$. Let $d_1 \ldots d_n$ be the results for this query.
   2.2. Let $i = 1$, $p_q = p$.
   2.3. While $p_q > 0$
      a. If $obsRel(d_i, q) > r$ Then
         If $obsRel(d_{i+1}, q) > obsRel(d_i, q) + c$
         Go to step (c)
         Otherwise
         Click on $d_i$.
         $p_q = p_q - 0.5 - (1 - rel(d_i, q))$.
         If $rel(d_i, q) = 1$ the user is done.
      b. Otherwise
         $p_q = p_q - (r - obsRel(d_i, q))$
      c. $i = i + 1$.
   2.4. With 50% probability, the user gives up.

more common words than others (for example consider two topics, basketball and machine learning). This construct is illustrated in Figure 3. In our experiments, each word is on average in two topics.

Next, we generate each document $d$ with $L$ words one at a time. We pick $k_d$, which specifies how many different topics $d$ is relevant to, as described in Table 1. Topics are picked according to a Zipf law to account for some topics being much more popular than others (again consider basketball versus machine learning). We set the relevance of the document to each topic to be proportional to the number of times the topic was picked with the sum of the relevances normalized to 1.

### 3.2. User Model

The process each user goes through as they search the web is specified in Table 2. This is a simple model, but as we will show it is reasonable and useful. Assume the user has a question $q$ and wants to find the most relevant documents to the related topic $T_q \in T$. Users differ in their patience $p$ and relevance threshold $r$. The patience determines how many results the user is likely to look at, while the relevance threshold specifies how relevant a document must appear to be (according to the abstract shown by the search engine) before the user clicks on it.

Given a question, the user generates a query. We implement this by sampling words from the question topic with a Zipf law. This query returns a set of results and the user considers each in order. When the user observes a result, she estimates it's relevance to her question given a short abstract, observing $obsRel(d_i, q)$. The real relevance of $d_i$ to query $q$ is $rel(d_i, q)$. $obsRel(d_i, q)$ is drawn from an incomplete Beta distribution with $\alpha$ dependent on the level of noise and $\beta$ selected so that the mode is at $rel(d_i, q)$ (unless $rel(d_i, q) = 0$, when the mode is at 0.05) as shown in Figure 4. This ensures the observed relevance is in the range [0,1] and has a level of noise that can be controlled.

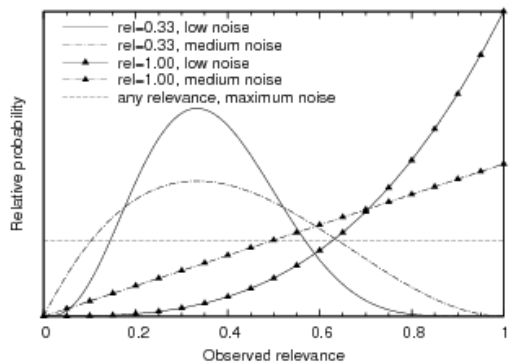

*Figure 4.* Probability of observing different perceived relevance as a function of the actual relevance.

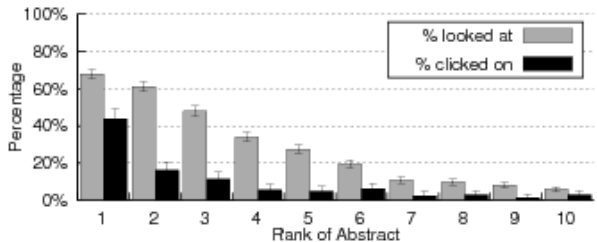

*Figure 5.* Percentage of time an abstract was viewed and clicked on depending on the rank of the result.

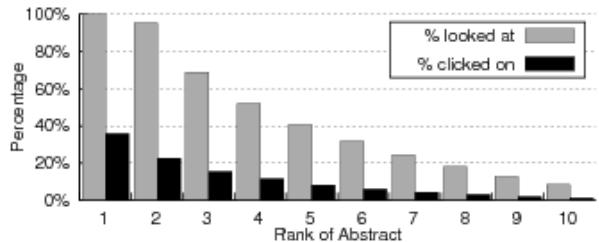

*Figure 6.* Percentage of time an abstract was viewed and clicked on in model depending on the rank of the result.

If $obsRel(d_i, q) > r$, the user's relevance threshold, the user intends to click on $d_i$. However, the eye tracking study described below showed that users typically look at the next document below any they click on. Hence before clicking, the user looks at the next document, and moves on to it if it appears substantially more relevant. Otherwise, if $obsRel(d_i, q) \leq r$, the user moves on and her patience is reduced. The patience is reduced more for documents that appear less relevant because if she sees a document that appears to be completely irrelevant, she is more discouraged than if she sees a document that appears somewhat relevant.

When the user clicks on a document, she sees $rel(d_i, q)$. If she finds a document with maximum relevance, she stops searching. Otherwise, she returns to the search results and continues looking until her patience runs out, and then runs a new query with 50% probability.

### 3.3. Model Justification

We base our usage model on results obtained in an eye tracking study (Granka, 2004; Granka et al., 2004; Joachims et al., 2005). The study aimed to observe how users formulate queries, assess the results returned by the search engine and select the links they click on. Thirty six student volunteers were asked to search for the answers to ten queries. The subjects were asked to start from the Google search page and find the answers. There were no restrictions on what queries they may choose, how and when to reformulate queries, or which links to follow. All clicks and the results returned by Google were recorded by an HTTP proxy. Movement of the eyes was recorded using a commercial eye tracker. Details of the study are provided in (Granka et al., 2004).

Figure 5 shows the fraction of time users looked at,

*Table 3.* Behavioral dimensions explored

| Short Name | Description |
|---|---|
| noise | Accuracy of relevance estimates. |
| ambiguity | Topic and word ambiguity. |
| trust | User's trust in presented ranking. |
| threshold | User selectivity over results. |
| patience | Number of results looked at. |
| reformulation | How often users reformulate. |
| improvement | Query improvement over time. |

and clicked on, each of the top 10 search results after running a query. It tells us that users usually look at the top two result abstracts, and are much more likely to click on the first result than any other. Additionally, (Joachims et al., 2005) show that users usually look sequentially at the results from the top to the one below the last one clicked on.

We observe in Figure 6 that the looks and clicks generated by this model resemble those seen in the user study. The most significant difference is in where users looked. Some of the time in the eye tracking study, the results show that users did not look at any results. We believe that this is partly due to errors in the eye tracker, and partly due to queries that did not return any results (such as spelling errors). For simplicity, we ignore these cases here.

We also measured the fraction of users who click on each of the top ten results in the Cornell University library search engine. The results confirmed that the distribution of clicks seen in Figures 5 and 6 is typical.

## 4. Learning Experiments

In this section, we explore the effect of different aspects of user behavior on the performance of Osmot. There are a number dimensions along which we assume user behavior may vary. These are listed in Table 3. For each, we present the effect of a change on our learning results and draw conclusions. Where possible, we relate the modeled results to real-world results to verify that the modeled results are realistic.

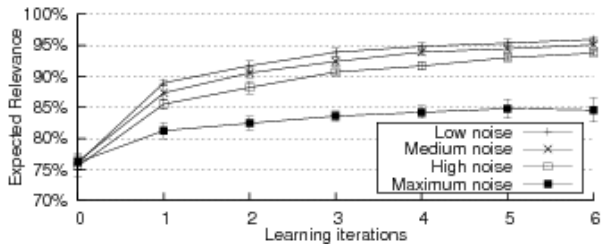

*Figure 7.* Ranking function performance for various noise levels.

### 4.1. High Level Parameters

We first consider the effect of two high level parameters: the level of difficulty users have in obtaining accurate relevance estimates from result abstracts, and the ambiguity in words appearing in documents and queries.

#### 4.1.1. Accuracy of relevance estimates

After running a query, users select where to click by estimating the relevance of results from the abstracts presented. We now vary the noise in the user relevance estimate and examine the effect.

Figure 7 shows the mean relevance of the most relevant document in the top five results for various noise levels for the first query in each query chain. This relevance is known because the evaluation is on a synthetic dataset. Consider the first set of points, at iteration 0. We used $rel_0$ as a ranking function and modeled 4,000 users running queries and clicking on results. This gave about 75% mean highest top-5 relevance. Each curve shows the performance of the learning algorithm for different levels of noise in users' estimates of document relevance. For each noise level, using the data generated we learned a new ranking function. These results are shown at iteration 1. We see that in each case performance improves and this improvement is smaller with more noise.

Using the learned ranking function, we collect more training data. We then use the training data to learn a second ranking function, re-evaluate (the results are shown at iteration 2) and so forth. The noise levels correspond to setting $\alpha$ to 4, 2, 1.4 and 1 in the incomplete Beta distribution.

We see that most of the improvement occurs in the first two learning iterations, although it keeps accruing. We also see that the decay in improvement as more noise is introduced is gradual, which tells us that the Osmot algorithm can be decays gracefully with more noise.

Given that the preferences are generated over a known document collection, we can measure the error in the constraints generated according to the real document relevance. In this analysis, we ignore all preferences that indicate a preference over documents that have the same true relevance to a query. The fraction of constraints indicating that two documents should be ranked one way while the reverse is true for the four noise levels considered are 5%, 11%, 18% and 48%. These percentages show the mismatch between the generated preferences and the known ground truth on the 0th iteration. They measure how often a preference indicates that $d_i >_q d_j$ when $d_i <_q^* d_j$ in reality.

In order to measure the level of noise in real data, we collected explicit relevance judgments for the data recorded during the eye tracking study. Five judges were asked to (weakly) order all result documents encountered during each query chain according to their relevance to the question (Radlinski & Joachims, 2005). From this data, we found that the inter-judge disagreement in real preference constraints generated according to Figure 1 is about 14%. Note that this is a different measure than above because we are comparing the preferences of two judges rather than preferences of one judge to a ground truth. This means that the error rate between the ground truth and a human judge is in the range 7-14%, depending on the level of independence between the judgments of the two judges. These results tell us that the error rate in the medium noise setting is likely to be realistic.

The maximum noise case is special because in this case the users effectively ignore the document abstracts when deciding whether to click. Despite this, we still observe improved performance as we run the learning algorithm. How can this be explained? As mentioned above, the error rate in these constraints is 48%, meaning that 52% of the constraints correctly state a valid preference over documents. This comes about because users still start from the top result and stop searching after finding (clicking on) a completely relevant document, producing some bias. Also note that we generate the most preferences for the last (and often completely relevant) document clicked on within a query chain. While some of this effect may be an artifact of our setup, we still find it interesting that this learning approach appears to be effective with such a small signal to noise ratio.

#### 4.1.2. Topic and word ambiguity

In the dataset used above, each word is on average in two topics. We also created collections where words were never in more than one topic, and where each word is on average in three topics. Figure 8 shows the results for the three collections. We see that with unambiguous words the ranking algorithm learns faster

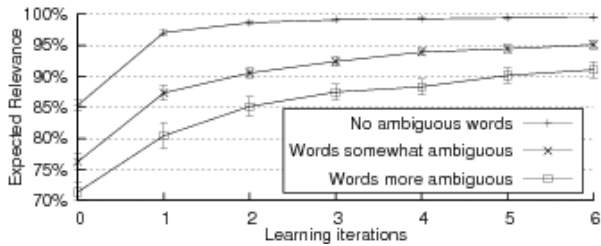

Figure 8. Ranking function performance for document collections with different levels of word ambiguity.

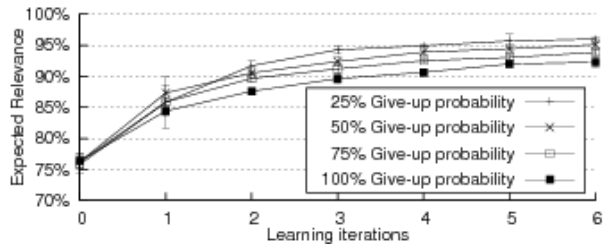

Figure 10. Ranking function performance for various probabilities that unsuccessful users will reformulate their query.

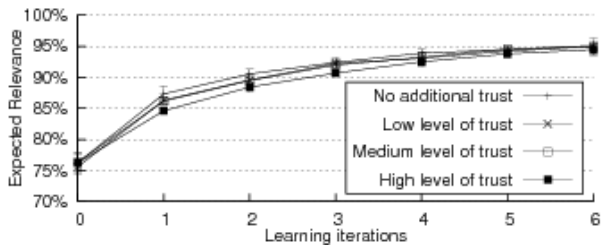

Figure 9. Ranking function performance versus the additional trust users place in the search engine.

and that even with more word ambiguity, our learning algorithm performs well.

### 4.2. Lower Level Parameters

The remainder of the behavioral dimensions are at a lower level, determining individual user behavior. We next explore the effect of these parameters.

#### 4.2.1. User trust in ranking presented

We saw earlier that users click surprisingly often on the top link. In fact, users appear to have inherent trust in Google that is not correlated to the relevance of the result abstracts (Joachims et al., 2005). We tested if such trust affects Osmot. Figure 9 shows that additional trust (implemented by increasing *obsRel* proportionally to the inverse of the rank of each result) has no lasting effect. This is interesting because it demonstrates that even when click-through feedback is strongly biased, it still provides useful training data.

An alternative explanation for users clicking predominantly on the top few results is that some users are more selective than others. Many may click on the first partially relevant result, i.e. the top one while others may only click on results that appear highly relevant. To test this, we added a constant to the threshold value picked in the user model. We found that performance was very similar over a reasonable range of values.

#### 4.2.2. Number of results looked at

Figure 5 also showed us that users look at surprisingly few of the search results. In order to explore the effect of this on the effectiveness of our learning approach, we changed the range of patience levels that users have. In the four settings tested, about 3%, 7%, 15% and 23% of users looked past the top 5 abstracts. The results showed that this has no significant effect on the performance for the first few iterations of learning, although the improvement in expected relevance tapers out faster in the case where users view fewer results. We omit the full results due to space constraints.

#### 4.2.3. How, and how often users reformulate

Previous work studying web search behavior (Lau & Horvitz, 1999; Silverstein et al., 1998) observed that users rarely run only one query and immediately find suitable results. Rather, users tend to perform a sequence of queries. Such query chains are also observed in the eye tracking study and our real-world search engine. Given Osmot's dependence on query chains, we wished to measure the effect of the probability of reformulation on the ranking function performance. The results are shown in Figure 10.

We see that the reformulation probability has a small but visible effect on ranking performance. While these results agree with our real-world experience that the presence of query chains makes a difference in algorithm performance (Radlinski & Joachims, 2005), we conjecture that in practice the difference is larger than seen here. In particular, unlike the model of user behavior presented in this paper, we suspect that later queries are not identically distributed to earlier queries. Rather we hypothesize that later queries are better and that this accounts for an additional improvement in performance when users chain multiple queries.

Using the relevance judgments of the five judges on the data gathered in the eye tracking study, we tested this hypothesis. Indeed, when a strict preference judgment is made by a human judge comparing the top result of two queries in a query chain, 70% of the time the top result of the later query is judged more relevant. We see a similar result when comparing the second ranked documents. We attempted to add such an effect to our model by making later queries progressively longer, but this did not end up having any discernible effect. We intend to explore this question more in the future.

## 5. Conclusions and Future Work

In this paper we have presented a simple model for simulating user behavior in a web search setting. We used this model to study the robustness of an algorithm for learning to rank that we previously found to be effective in a real-world search engine. We demonstrated that the learning method is robust to noise in user behavior for a number of document collections with different levels of word ambiguity. Our results are important because they show that modeling allows fast explorations of the properties of algorithms for learning to rank. Although a more realistic model of user search behavior can be constructed, we have presented a reasonable starting model.

The model currently has a number of limitations that we intend to improve upon in the future. However, we believe that even in its present state it provides a valuable tool for understanding the performance of algorithms for learning to rank. We plan to make our implementation available to the research community.

## 6. Acknowledgments

We would like to thank Laura Granka, Bing Pang, Helene Hembrooke and Geri Gay for their collaboration in the eye tracking study. We also thank the subjects of the eye tracking study and the relevance judges. This work was funded under NSF CAREER Award IIS-0237381 and the KD-D grant.